\documentclass[conference,a4paper]{IEEEtran}

\IEEEoverridecommandlockouts

\usepackage[T1]{fontenc}
\usepackage[utf8]{inputenc}
\usepackage{graphicx}
\usepackage{booktabs}
\usepackage{array}
\usepackage{cite}
\usepackage{amsmath}
\usepackage{textcomp}
\usepackage{url}

\newcommand{\urtccopyright}{979-8-3195-2794-3/26/\$31.00~\textcopyright{}2026 IEEE}

\title{EEGBiFormer: Real-Time EEG Attention Classification via Hybrid CNN--Transformer with EEG-Informed Auditory Protocol Validation}

\author{
\IEEEauthorblockN{Isabella Guan}
\IEEEauthorblockA{\textit{Lake Washington High School}\\
Kirkland, WA, USA\\
isabellaguan2009@gmail.com}
}

\begin{document}

\IEEEpubid{\makebox[\columnwidth]{\urtccopyright\hfill}\hspace{\columnsep}\makebox[\columnwidth]{}}

\maketitle

\begin{abstract}
Sustained attention is critical for academic performance, yet existing auditory support tools are static and not grounded in real-time neural evidence. This paper presents EEGBiFormer, a hybrid CNN--Transformer model for real-time binary classification of attention states, focus versus non-focus, from 16-channel EEG signals. A custom analog front-end acquisition board was designed and built to collect a task-based EEG dataset. Before model training, spectral and connectivity analysis, using Power Spectral Density and Phase Locking Value, confirmed distinct neural signatures between attention states, providing physiological grounding for the classification approach. EEGBiFormer achieved 91.3 percent window-level validation accuracy with an F1 score of 94.5 percent, and 88.2 percent recording-level accuracy across 17 independent sessions. End-to-end system latency was measured below 100 ms, supporting real-time responsiveness. The trained model was further used to objectively rank auditory stimuli by focus-state confidence, yielding EEG-informed audio protocols. In a paired within-subject behavioral study with 37 participants, these protocols produced a statistically significant 20 percent improvement in composite attention scores over silent controls, with a medium-to-large effect size. This work demonstrates a complete pipeline from custom EEG hardware through neural signal analysis, deep learning classification, and evidence-based auditory protocol design.
\end{abstract}

\begin{IEEEkeywords}
EEG, attention classification, CNN--Transformer, binaural beats, real-time classification, biomedical engineering
\end{IEEEkeywords}

\section{Introduction}

Sustained attention is essential for effective learning, yet many adolescents experience significant difficulties sustaining attention during academic tasks, particularly those with Attention-Deficit/Hyperactivity Disorder (ADHD). Auditory strategies such as background music, white noise, and binaural beats (BB) are widely used informally to support focus, with therapeutic benefits for attention and cognitive performance reported particularly in clinical and autism-spectrum populations~\cite{geng2023,zhu2021,sanyal2018,sharda2018,kang2021}. In the general adolescent population, however, such strategies are typically selected by subjective preference rather than objective neural evidence.

Electroencephalography (EEG) provides a non-invasive window into cortical dynamics at millisecond resolution~\cite{gurau2017,li2019}. Prior research has identified reliable neural correlations of attentional engagement: elevated theta activity (4--8 Hz) and reduced alpha power (8--13 Hz, ``alpha blocking'') are consistently observed during focused cognitive states, while inter-regional phase synchrony, quantified via Phase Locking Value (PLV), reflects the large-scale functional network coordination that underlies sustained attention~\cite{shou2023,wang2022,provorova2024,freeman2002}.

\IEEEpubidadjcol

Despite this neuroscientific foundation, existing attention-support audio tools remain disconnected from physiological data. They neither adapt to the user's real-time neural state nor use objective EEG measures to guide audio selection. This gap motivates the present work.

The work consists of three main contributions: (1) a custom 16-channel EEG acquisition system and task-based dataset; (2) EEGBiFormer, a lightweight hybrid Convolutional Neural Network (CNN)--Transformer achieving 91.3\% accuracy for binary attention classification; and (3) an EEG-informed audio selection methodology validated in a 37-participant behavioral study ($p < 0.001$).

\section{Hardware and Data Collection}

\begin{figure}[t]
    \centering
    \includegraphics[width=\columnwidth]{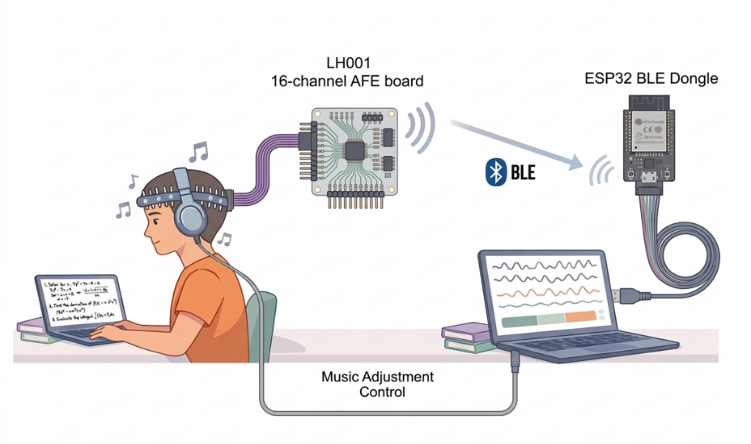}
    \caption{EEGBiFormer real-time pipeline.}
    \label{fig:pipeline}
\end{figure}

Fig.~\ref{fig:pipeline} shows the target closed-loop architecture. A custom 16-channel EEG headset streams 202 Hz data over Bluetooth Low Energy to a host PC, where EEGBiFormer classifies focus state and a decision-logic module selects an audio track with end-to-end latency under 100 ms (Table~\ref{tab:hardware}). Because continuous live operation was impractical for a multi-participant study, the pipeline was validated in two stages: the model first ranked candidate audio stimuli by focus-state confidence (Section~\ref{sec:audio}), then the top-ranked stimuli were delivered as fixed audio in a 37-participant paired study (Section~\ref{sec:behavior}).

\subsection{Custom EEG Acquisition System}
\label{sec:hardware}

To collect proprietary EEG data without relying on commercial headsets, a custom 16-channel acquisition board was designed using EasyEDA and manufactured through JLCPCB. Two daisy-chained LH001-99 AFE chips sample all 16 channels at 202 Hz ($<0.5~\mathrm{\mu V}$ RMS noise), while an ESP32-S3 handles SPI acquisition and streams data over BLE, keeping end-to-end latency under 100 ms. Multi-channel EEG enables spatial discrimination of synchronous neural sources across cortical regions~\cite{liu2006}.

The electrode interface went through two iterations. An initial wet Ag/AgCl gel design suffered rising impedance ($5~\mathrm{k\Omega}\rightarrow >50~\mathrm{k\Omega}$) as the gel dried, yielding usable data in only 6 of 20 pilot sessions and capping accuracy at 63\% despite software mitigations. Switching to dry stainless-steel comb electrodes traded a higher but stable impedance ($20$--$50~\mathrm{k\Omega}$) for consistent signal quality across a full session, improving classification accuracy by 11.2 percentage points with no model changes. Table~\ref{tab:hardware} summarizes the final hardware configuration used for all reported results.

\begin{table}[t]
\caption{Hardware Component Specifications}
\label{tab:hardware}
\centering
\footnotesize
\begin{tabular}{@{}>{\raggedright\arraybackslash}p{0.29\columnwidth}>{\raggedright\arraybackslash}p{0.64\columnwidth}@{}}
\toprule
\textbf{Component} & \textbf{Specification} \\
\midrule
Analog Front-End & 2$\times$ LH001-99 AFE chips, daisy-chain (16 ch) \\
Sampling Rate & 202 Hz per channel \\
System Noise & $<0.5~\mathrm{\mu V}$ RMS \\
Electrode Type & Comb electrodes \\
Wireless & Bluetooth Low Energy (BLE), ESP32-S3 \\
Battery & 3.7 V Li-polymer, $\geq 4$ h continuous \\
PCB & Separate analog/digital ground planes \\
End-to-End Latency & $<100$ ms (BLE acquisition $\rightarrow$ inference $\rightarrow$ output) \\
\bottomrule
\end{tabular}
\end{table}

\begin{figure}[t]
    \centering
    \includegraphics[width=\columnwidth]{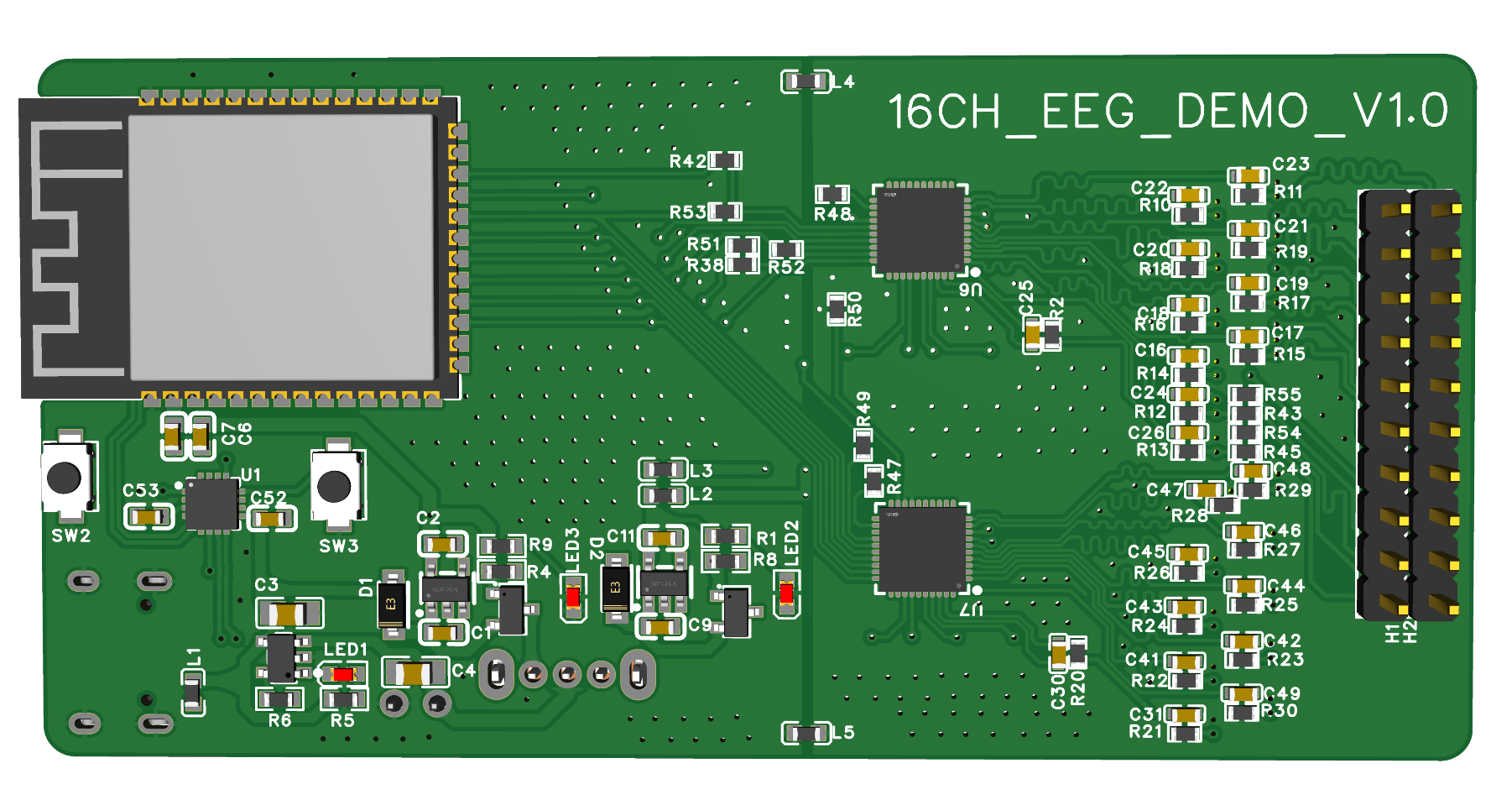}
    \caption{Custom 16CH\_EEG\_DEMO\_V1.0 acquisition board.}
    \label{fig:pcb}
\end{figure}

\subsection{Recording Paradigm and Dataset}
\label{sec:dataset}

EEG data were collected using a task-based paradigm. Participants performed structured attention-demanding tasks (reading, symbol identification) interleaved with controlled low-engagement intervals to produce labeled focus and non-focus segments. A sliding window approach (2 s window, 50\% overlap, 202 Hz $\rightarrow$ 404 samples/channel) was applied to generate fixed-length samples. Session-level separation was maintained between splits to prevent data leakage.

\begin{table}[t]
\caption{Dataset Split by Session}
\label{tab:dataset}
\centering
\footnotesize
\begin{tabular}{@{}lrrr@{}}
\toprule
\textbf{Split} & \textbf{Focus} & \textbf{Non-focus} & \textbf{Total} \\
\midrule
Training & 27 & 11 & 38 \\
Validation & 8 & 3 & 11 \\
\bottomrule
\end{tabular}
\end{table}

\section{EEG Signal Analysis: PSD and PLV}

Before model training, exploratory spectral and connectivity analysis was conducted to confirm that the dataset contains discriminable neural signatures. These analyses also motivated the selection of 40 Hz binaural beats as a candidate auditory intervention.

\subsection{Power Spectral Density Analysis}

Power Spectral Density (PSD) was computed across all 16 channels for focus and non-focus segments. Topographic distribution maps show the scalp-level power distribution across the three attention-relevant bands.

\begin{figure}[t]
    \centering
    \includegraphics[width=\columnwidth]{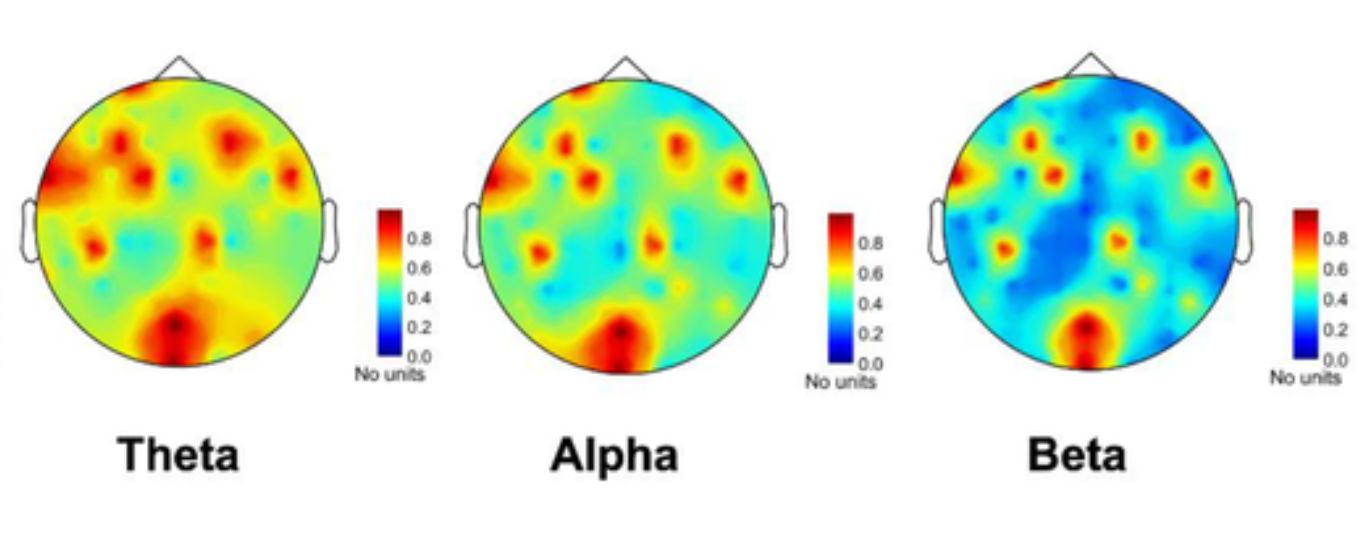}
    \caption{PSD topographic maps (Theta 4--8 Hz, Alpha 8--13 Hz, Beta 14--30 Hz). Warm colors indicate higher relative power.}
    \label{fig:psd}
\end{figure}

Consistent with established neuroscience literature, focus sessions showed suppressed frontal alpha power (alpha blocking) and elevated parietal-occipital beta activity relative to non-focus sessions. These spectral differences confirm that the task-based paradigm successfully induced distinguishable attentional states.

PSD analysis was additionally used to evaluate the neural response to 40 Hz binaural beats, a stimulus frequency in the gamma band associated with cognitive binding. Channel-averaged PSD plots for focus and non-focus conditions during BB exposure were compared, with a reference marker at the 40 Hz target frequency~\cite{battu2025}.

\subsection{Phase Locking Value Analysis}

\begin{figure}[t]
    \centering
    \includegraphics[width=\columnwidth]{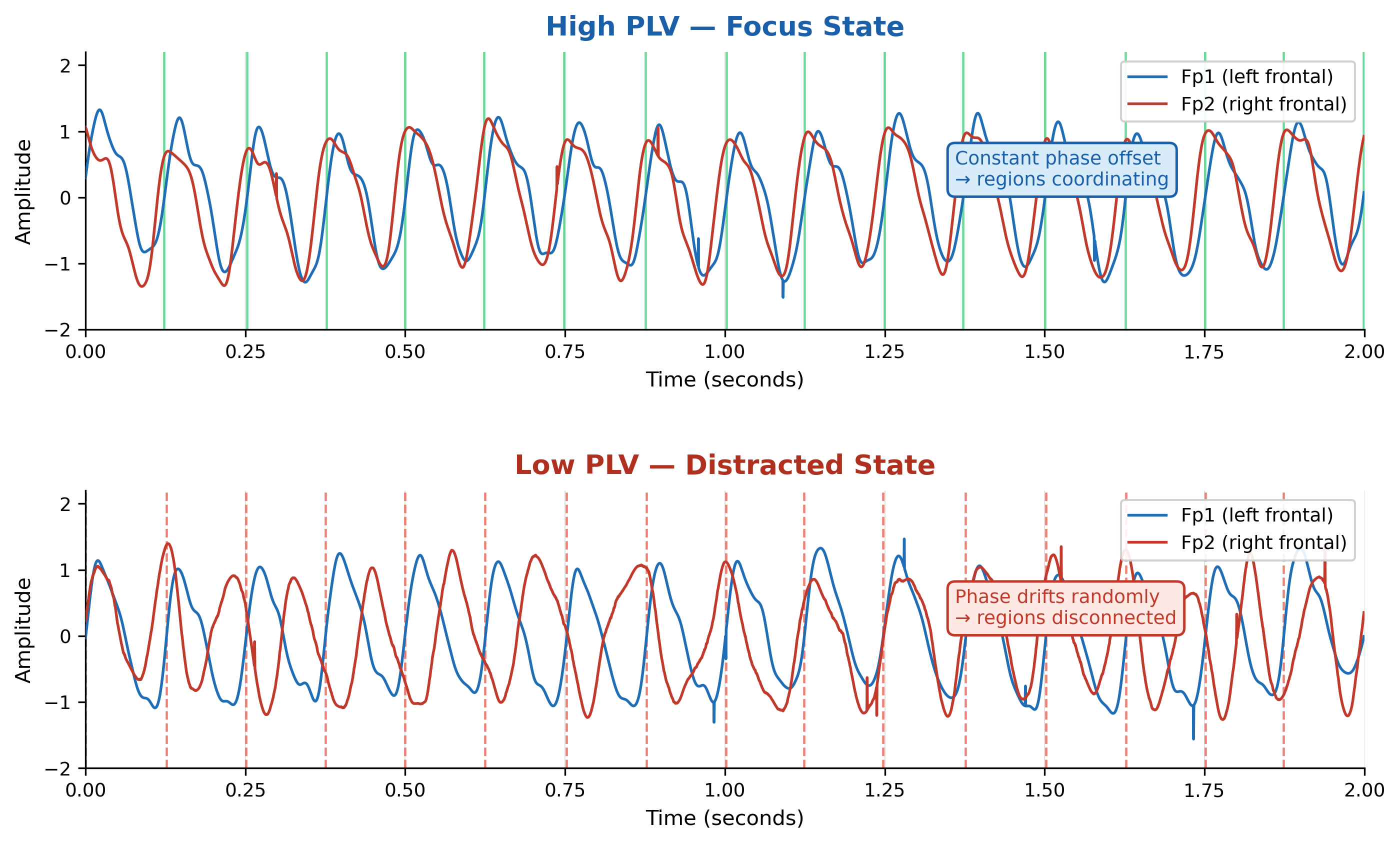}
    \caption{PLV analysis (Fp1--Fp2, 40 Hz). Focus sessions show stable phase offset (high PLV); distracted sessions show phase drift (low PLV).}
    \label{fig:plv}
\end{figure}

PLV measures the consistency of the phase difference between two EEG channels across time, quantifying inter-regional functional connectivity~\cite{provorova2024,freeman2002}. PLV ranges from 0 (no synchrony) to 1 (perfect phase locking).

During focus sessions, Fp1 and Fp2 maintain a constant phase relationship, indicating coordinated inter-hemispheric frontal activity. During distraction, phase drifts randomly, reflecting reduced functional connectivity. This finding supports PLV as a meaningful attention biomarker and provides neurophysiological justification for targeting 40 Hz in the auditory intervention design.

\subsection{EEG-Informed Audio Selection}
\label{sec:audio}

The trained EEGBiFormer model was applied to EEG recordings collected while participants listened to six candidate audio stimuli. The mean focus-state classification confidence was computed for each stimulus type and used to rank their neural effectiveness objectively.

Nature sounds (Ocean Waves: 94.2\%; Rain Falling: 93.5\%) and 40 Hz binaural beats combined with classical music (F\"ur Elise BB: 92.8\%) ranked highest. The silent baseline (81.3\%) confirmed that all tested stimuli provided measurable neural benefit. The top-ranked stimuli were used as the EEG-informed audio protocol in the behavioral validation study (Section~\ref{sec:behavior}).

\begin{figure}[t]
    \centering
    \includegraphics[width=\columnwidth]{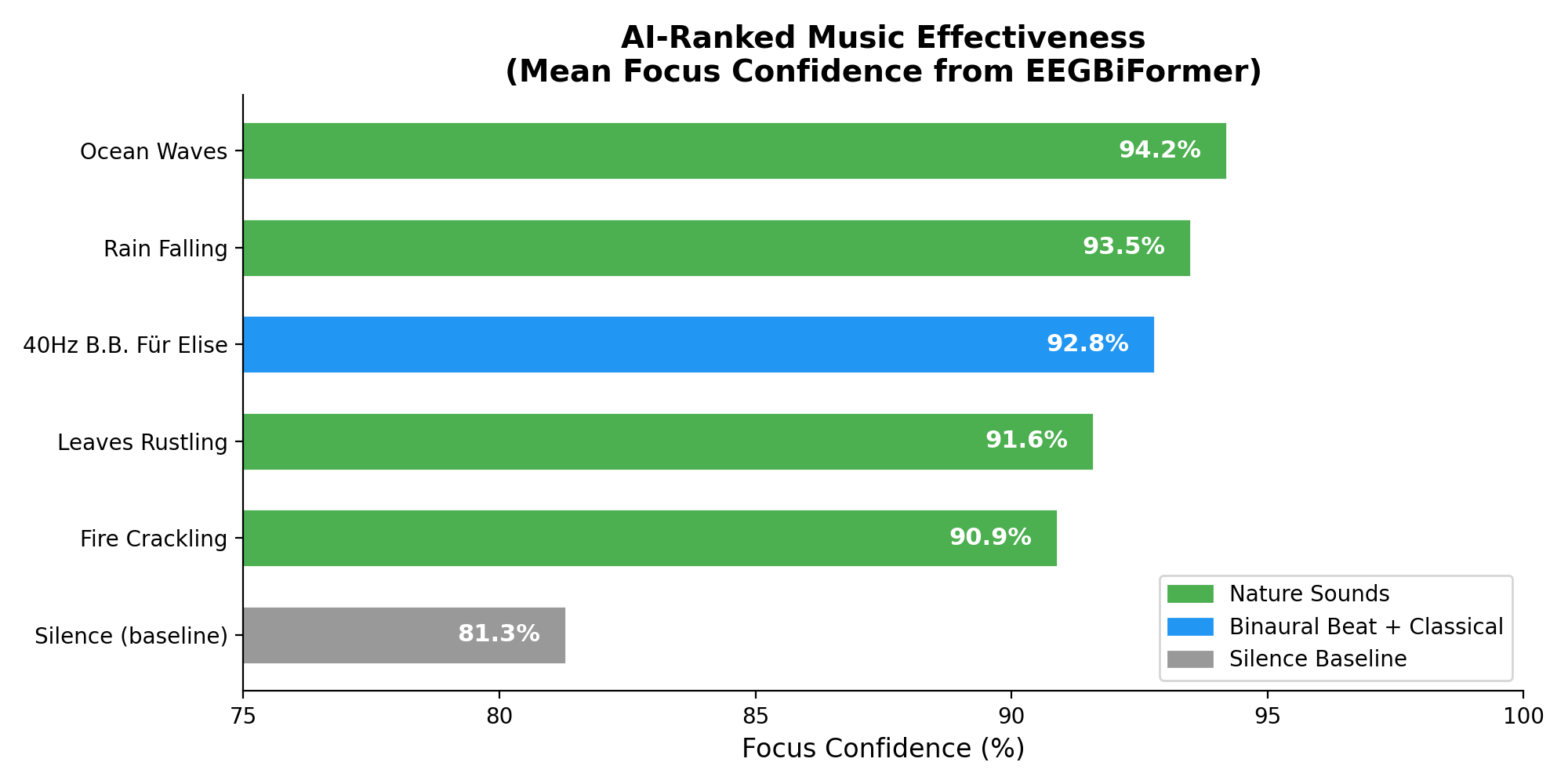}
    \caption{EEG-informed audio ranking.}
    \label{fig:ranking}
\end{figure}

\section{EEGBiFormer: Hybrid CNN--Transformer Architecture}

\subsection{Model Design}

EEGBiFormer processes raw 16-channel EEG windows (shape: $404\times16$) through a two-stage hybrid architecture.

\textbf{CNN Front-End---}Three stacked Conv1D layers extract spatiotemporal features directly from multi-channel raw signals~\cite{lawhern2018}. A 9-sample kernel ($\approx44.5$ ms at 202 Hz) spans one Beta-band cycle and half an Alpha-band cycle. Stride-1, same-length convolutions preserve the 404-sample sequence while channel depth expands $16\rightarrow64\rightarrow128\rightarrow256$. Stacking the layers expands the effective temporal receptive field to approximately 25 samples ($\approx124$ ms). No pooling is applied, so the full $404\times256$ sequence remains available to the Transformer.

\begin{figure}[t]
    \centering
    \includegraphics[width=\columnwidth,height=0.46\textheight,keepaspectratio]{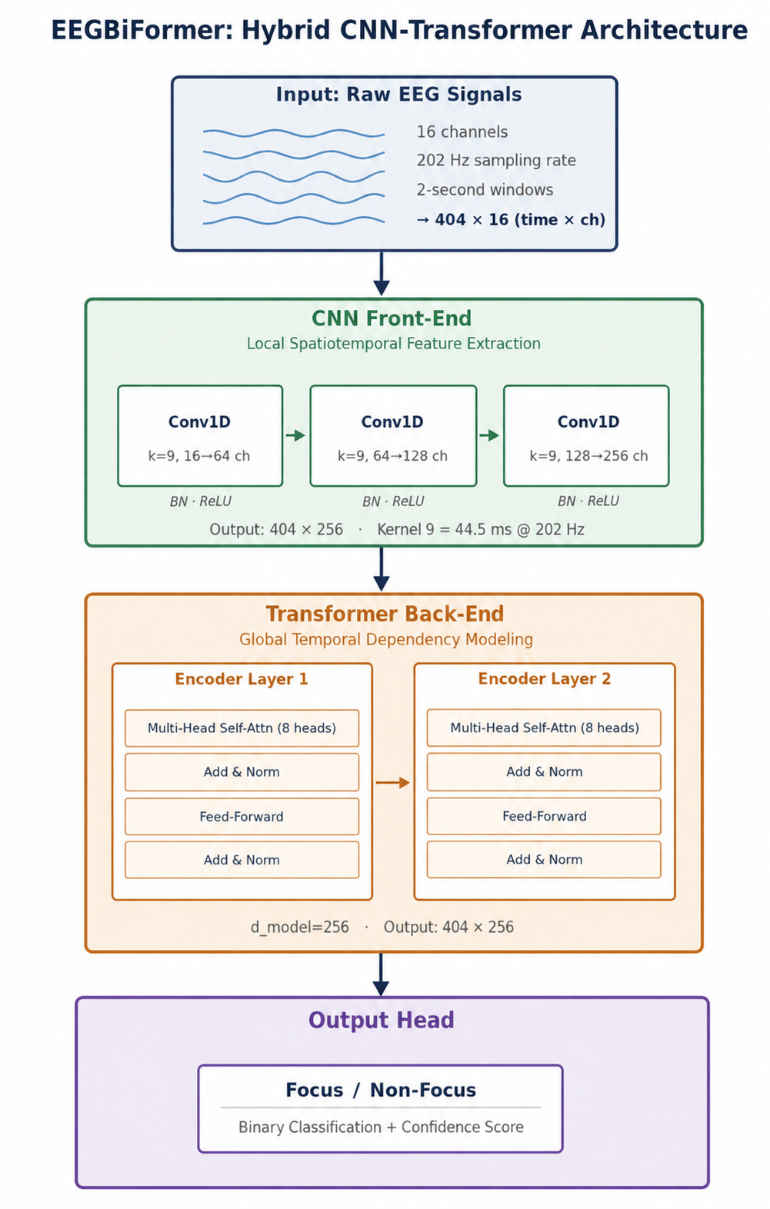}
    \caption{EEGBiFormer architecture: the CNN front-end extracts local features; the Transformer back-end models global temporal dependencies.}
    \label{fig:architecture}
\end{figure}

\textbf{Transformer Back-End---}Two independently trained encoder layers use 8-head self-attention to model long-range dependencies within each 2-second window. Every time step can attend to all others, allowing the heads to learn patterns at different temporal scales. The first layer models dependencies in the CNN features, and the second models relationships in that contextualized output. Two layers balance representational depth against overfitting risk on the limited training set.

\textbf{Output Head---}Global Average Pooling condenses the $404\times256$ output into a 256-dimensional vector. A Linear layer ($256\rightarrow2$) with dropout ($p=0.2$) produces Focus/Non-Focus logits; softmax converts them to class probabilities and the larger value is reported as confidence.

\begin{table}[t]
\caption{EEGBiFormer Model Hyperparameters}
\label{tab:hyperparameters}
\centering
\scriptsize
\begin{tabular}{@{}p{0.17\columnwidth}p{0.17\columnwidth}p{0.16\columnwidth}p{0.32\columnwidth}@{}}
\toprule
\textbf{Component} & \textbf{Parameter} & \textbf{Value} & \textbf{Rationale} \\
\midrule
CNN & Layers & 3 & Hierarchical feature extraction \\
CNN & Kernel size & 9 & Matches Beta/Gamma oscillation periods \\
CNN & Channel depth & $16\!\rightarrow\!64\!\rightarrow\!128\!\rightarrow\!256$ & Progressive representation capacity \\
Transformer & Encoder layers & 2 & Models first- and second-order dependencies \\
Transformer & Attention heads & 8 & Parallel multi-scale temporal modeling \\
Output & Pooling & Global Average & Sequence aggregation, parameter efficiency \\
Regularization & Dropout & 0.2 & Prevents overfitting \\
\bottomrule
\end{tabular}
\end{table}

\subsection{Training Strategy}

Three techniques addressed the challenges of limited and imbalanced training data:

\textbf{Weighted cross-entropy loss:} The training set contains 2,244 focus windows and 540 non-focus windows (ratio $\approx4.1{:}1$). Class weights were computed inversely proportional to frequency (focus: 0.62; non-focus: 2.58), penalizing minority-class errors more heavily.

\textbf{AdamW optimizer:} Decoupled weight decay regularization was applied to all parameters, correcting the under-regularization of the standard Adam optimizer when combined with L2 penalties.

\textbf{ReduceLROnPlateau scheduler:} The learning rate was reduced by $\times0.1$ when validation loss failed to improve for 15 consecutive epochs, enabling fine-grained convergence near the optimum.

\section{Results}

\subsection{Classification Performance}

On the held-out validation set (window-level evaluation), EEGBiFormer achieved the classification performance summarized in Table~\ref{tab:validation}. The corresponding confusion matrix comprised TP = 2,073, FN = 171, TN = 469, and FP = 71, totaling 2,784 evaluated windows, substantially more than the number of underlying recordings, since each validation recording is segmented into multiple 2-second, 50\%-overlap windows via the sliding-window procedure described in Section~\ref{sec:dataset}.

Training loss decreased from 0.68 to approximately 0.05, with training accuracy approaching 100\%. Validation loss and accuracy tracked closely alongside these curves without diverging, indicating stable convergence with minimal overfitting. The final reported model (91.3\% validation accuracy) reflects continued optimization beyond this representative run.

\begin{table}[t]
\caption{Validation Set Performance}
\label{tab:validation}
\centering
\begin{tabular}{@{}lr@{}}
\toprule
\textbf{Metric} & \textbf{Value} \\
\midrule
Accuracy & 91.3\% \\
Precision (Focus) & 96.7\% \\
Recall (Focus) & 92.4\% \\
F1 Score & 94.5\% \\
\bottomrule
\end{tabular}
\end{table}

\subsection{Cross-Session Generalization}

To assess out-of-distribution robustness, the model was evaluated on an independent offline dataset ($n=17$ recordings) collected during naturalistic activities (reading, drawing, social interaction). Unlike the window-level evaluation in Section~\ref{sec:dataset}, evaluation here was performed at the recording level: each recording's final label is determined by majority vote across all of its constituent 2-second windows.

\begin{table}[t]
\caption{Cross-Session Performance}
\label{tab:cross-session}
\centering
\begin{tabular}{@{}lr@{}}
\toprule
\textbf{Metric} & \textbf{Value} \\
\midrule
Recording-level Accuracy & 88.2\% \\
Precision (Focus) & 90.9\% \\
Recall (Focus) & 90.9\% \\
F1 Score & 90.9\% \\
\bottomrule
\end{tabular}
\end{table}

At the recording level ($n=17$), the confusion matrix was TP = 10, FN = 1, FP = 1, and TN = 5. These results held up qualitatively as well: a session of sustained Shakespeare reading was classified as ``focus'' with 99.7\% confidence (unanimous across windows), while a casual card game was classified as ``non-focus'' with 68.0\% confidence. This close agreement between window-level validation performance and recording-level offline performance (F1: 94.5\% vs. 90.9\%) suggests reasonable cross-session generalization despite the limited dataset size.

\subsection{Behavioral Validation of EEG-Informed Audio Protocols}
\label{sec:behavior}

To evaluate the downstream utility of the EEG-informed audio selection, a within-subject, counterbalanced paired behavioral study was conducted with 37 participants. The sample included both youth and adult participants, some of whom self-reported an ADHD diagnosis.

Each participant completed two 10-minute standardized testing reading/math sessions: an EEG-informed audio condition, featuring one of the top-5 EEG-ranked sounds (nature sounds or 40 Hz binaural beats), and a silent control. Condition order was counterbalanced (19 audio-first, 18 silence-first) to reduce fatigue and practice effects.

After each session, participants completed a self-report questionnaire. A Composite Attention Score was calculated as the mean of three items assessing focus, distraction, and mental engagement, then linearly rescaled from 1--5 to 0--100.

A paired $t$-test on the 37 within-subject differences yielded $t(36)=4.01$, $p<0.001$ (two-tailed), 95\% CI [6.5, 19.9], Cohen's $d=0.66$ (medium-large effect). The 20.0\% relative improvement, $(79.2-66.0)/66.0$, was consistent across individual questionnaire items: sustained attention, distraction resistance, and perceived engagement all showed directional improvement under the EEG-informed condition.

\begin{table}[t]
\caption{Behavioral Validation: Composite Attention Scores}
\label{tab:behavior}
\centering
\begin{tabular}{@{}lrrr@{}}
\toprule
\textbf{Condition} & \textbf{Mean} & \textbf{SD} & \textbf{Range} \\
\midrule
EEG-informed audio & 79.2 & 18.6 & 47--100 \\
Silence control & 66.0 & 21.2 & 33--93 \\
Difference & +13.2 & 20.0 & --- \\
\bottomrule
\end{tabular}
\end{table}

\section{Discussion and Conclusion}

This paper demonstrates a complete, evidence-based pipeline for EEG-informed attentional support. EEGBiFormer achieves strong binary attention classification (91.3\% validation accuracy, 88.2\% cross-session) from self-collected 16-channel EEG data, with end-to-end latency below 100 ms.

The gap between precision (96.7\%) and recall (92.4\%) reflects the weighted cross-entropy loss. Because non-focus windows are rarer (ratio $\approx4.1{:}1$), the loss penalizes their misclassification more heavily and favors predicting ``focus'' only at high confidence. This trades slightly lower recall for higher precision, which is desirable because triggering a focus-oriented intervention while the user is distracted is more disruptive than occasionally missing a focus state.

The PSD and PLV analyses serve a dual role: they validate the physiological quality of the dataset and provide interpretable neural grounding for the model's learned representations. The PLV finding that 40 Hz frontal synchrony distinguishes focus from distraction motivates the specific inclusion of 40 Hz binaural beats in the audio protocol, linking neuroscience directly to intervention design.

The behavioral validation ($p<0.001$, $d=0.66$) demonstrates that EEG-based audio ranking yields protocols meaningfully superior to unguided static alternatives. This bridges the gap between neural measurement and practical intervention, replacing subjective audio selection with objective physiological evidence.

\textbf{Limitations and Future Work---}The dataset is self-collected and limited in size ($n=49$ sessions), and results are specific to the custom acquisition hardware (Section~\ref{sec:hardware}); generalization to other headsets or electrode types remains untested. Binary classification does not capture the continuous spectrum of attention, and the behavioral study relies on self-report rather than objective performance metrics. The closed-loop system in Fig.~\ref{fig:pipeline} was validated for latency and technical feasibility, but the behavioral study delivered audio statically rather than via live EEG-triggered switching. Future work should expand participant diversity, explore continuous attention regression, validate in naturalistic classroom settings, and test the fully closed-loop system with human subjects.

\textbf{Conclusion---}EEGBiFormer establishes a technically feasible and behaviorally validated pathway from custom EEG hardware through neural analysis, deep learning classification, and evidence-based auditory protocol design, demonstrating how physiological signals can objectively inform cognitive support strategies.

\section*{Acknowledgment}

The author thanks Dr. Fusheng Wang and Ph.D. student Rui Liu, from the CSIRE research program of Stony Brook University, for their mentorship of this research.

\end{document}